\documentclass{article}

\usepackage[preprint]{neurips_2024}
\usepackage{graphicx}
\usepackage{caption}
\usepackage{float}
\usepackage{lmodern}
\usepackage[utf8]{inputenc} 
\usepackage[T1]{fontenc} 
\usepackage{xcolor}
\definecolor{darkblue}{rgb}{0.1, 0.2, 0.75}
\usepackage{xr}
\usepackage[hidelinks]{hyperref}
\hypersetup{
    colorlinks=true,
    linkcolor=darkblue, 
    citecolor=darkblue, 
    urlcolor=darkblue   
}
\usepackage{url}           
\usepackage{booktabs}       
\usepackage{amsfonts}       
\usepackage{nicefrac}      
\usepackage{microtype}      
\usepackage[dvipsnames]{xcolor}         
\usepackage{lineno}
\usepackage{tabularx}
\usepackage{makecell}
\usepackage{threeparttable}
\usepackage{amsmath}
\usepackage{footnote}

\usepackage{tikz}                  
\usepackage{cleveref}               
\newcommand{\added}[1]{\textcolor{black}{#1}}  
\usepackage{arydshln}       
\usepackage{subcaption}     
\usepackage{rotating}       
\usepackage{comment}        
\usepackage{nameref}    
\usepackage{lineno}     

\title{How AI writing assistance distorts perceptions of AI users and their opinions}

\author{Paul Röttger$^{1,2}$\thanks{Corresponding author: paul.rottger@hpi.de} , Kobi Hackenburg$^{2,3}$, Hannah Rose Kirk$^{2,3}$,\\ and \textbf{Christopher Summerfield}$^{2,3}$ \vspace{0.25cm} \\
$^{1}$Hasso Plattner Institute, $^{2}$University of Oxford, $^{3}$UK AI Security Institute}

\begin{document}

\maketitle 

\vspace{0.6cm}

\begin{abstract}
\added{
Hundreds of millions of people use AI for writing assistance.
We asked how AI distorts perceptions of AI users and their opinions.
Writers (N=2,939) wrote political opinion paragraphs with and without AI assistance.
Separate readers (N=11,091) blindly evaluated these paragraphs and made inferences about the writers.
With AI, writers seemed more opinionated, competent, positive, and demographically privileged.
Writers objected to many of these distortions yet preferred AI-assisted text even when made aware of them.
Reward models trained on our data mitigated objectionable distortions but reduced user acceptance.
Additional readers (N=8,798) trusted AI-assisted writers more and were more persuaded by AI writing when AI was more distortive.
Overall, distortions from AI writing assistance are pervasive, persistent under human oversight, and likely to have consequential effects on human behaviours and attitudes.
}
\end{abstract}

\newpage

\section*{Introduction}
\label{sec:intro}
\vspace{-0.4cm}

Written language is a primary medium through which people share information, understand one another, and reach agreement.
Today, artificial intelligence (AI) writing tools are transforming the writing process~\citep{hancock2020ai}:\
hundreds of millions of people now use AI to draft and refine text~\citep{chatterji2025people,handa2025economic,tomlinson2025working}, and AI-assisted writing already pervades \added{high-stakes settings in politics, law, academia, and industry}~\citep{harrington2023cnet,la2025machines,liang2025widespread,pimlico2025chatgpt,sun2025ai,wu2025federal,he2026academic,lee2026digital}.
\added{It is increasingly well documented that AI-assisted writing differs from texts produced by human writers alone because AI models tend towards particular word choices, tonal registers, and rhetorical patterns~\citep{anderson2024homogenization,padmakumar2024does,agarwal2025ai,jiang2025artificial,abdulhai2026llms,sourati2026shrinking}.
However, the downstream effects of these textual differences depend on how they are perceived:\
writing carries rich psychological signals~\citep{pennebaker2003psychological,ireland2014natural,kennedy2021moral,boyd2022development}, and readers routinely make inferences about writers and their beliefs based on the texts they encounter~\citep{walther1992interpersonal,hancock2001impression,flekova2016analyzing}.
These inferences play a key role in how readers relate to and are affected by writers~\citep{hovland1951influence,mackie1990processing,pornpitakpan2004persuasiveness,hohenstein2023artificial}.
Thus, in this article, we investigate \textit{distortions} from AI writing assistance not as textual changes but as \textit{systematic differences in how writers and their opinions are perceived, caused by their use of AI}.}

\added{These distortions} could have far-reaching consequences across the many domains where written communication shapes social life.
\added{Distortions of perceived political stance, for example,} could fuel misperceptions of public opinion, deepen partisan animosity, and reduce willingness to engage across ideological lines~\citep{westfall2015perceiving,levendusky2016mis,druckman2022mis,jost2022cognitive}.
\added{Distortions of writing quality or apparent expertise could lend additional persuasive power to AI-assisted arguments and foster undue trust~\citep{pornpitakpan2004persuasiveness, karmarkar2010believe,kumkale2010effects}.}
\added{Distortions of emotional tone could amplify intergroup hostility or suppress mobilization around genuine grievances~\citep{berger2012makes,brady2017emotion,rathje2021out}}.
\added{And distortions of a} writer's demographic background -- such as their perceived education, race, age, or gender -- could alter the personal and professional opportunities they are offered~\citep{bertrand2004emily,moss2012science,quillian2017meta}.
At the scale of billions of user requests for AI writing assistance~\citep{chatterji2025people}, \added{even modest distortions along these dimensions} could accumulate into \added{consequential effects for society}.

Critically, however, the extent and manner in which distortions from AI writing assistance will impact society depend on four open empirical questions.
First, does AI writing assistance actually distort how readers perceive writers and their opinions, and if so, when and in what ways (\textbf{RQ1})?
If distortions are negligible, occur along inconsequential dimensions, or vanish under realistic conditions where writers can freely edit and reject AI-generated text, real-world impacts may be marginal.
Second, if distortions occur, \added{how do writers feel about them} (\textbf{RQ2})?
If writers oppose the distortions that AI introduces, the normative concern is one of individual agency \added{and user experience}.
But if writers welcome distortions, the concern becomes collective \added{as distortions propagate}.
Thus, third, can undesirable distortions be mitigated without decreasing user preference for AI writing assistance (\textbf{RQ3})?
If targeted interventions at the model level can reduce specific distortions, developers have a tractable path to mitigating risks of AI writing assistance.
But if the textual properties that drive undesirable distortions are entangled with those that writers value, then some degree of distortion may be an inherent cost of AI writing assistance.
Fourth and finally, do distortions from AI writing assistance have consequential effects on human behaviours and attitudes (\textbf{RQ4})?
If AI use merely changes perceptions of writers and their opinions but has no further downstream consequences, then even pervasive distortions may be of limited practical concern.

Here, we address these questions using \added{five} large-scale studies, across which 2,939 writers \added{composed political} opinion paragraphs with and without writing assistance from one of three leading AI models.
Separate panels of 11,091 readers blindly evaluated both sets of paragraphs -- human-written and AI-assisted -- across 29 dimensions of reader perception, spanning political opinion, writing quality, writer personality, emotions, and demographics.
By comparing how the same writers were perceived through their own words versus AI-assisted text that writers had edited and approved, we measured the distortions that AI writing assistance introduces even under realistic conditions of human oversight.
In subsequent analyses, we measured writer tolerance for these distortions and tested whether informing writers of distortions changes their willingness to adopt AI-assisted text.
Using the dataset of 10,008 paragraphs with 2,903,596 ratings from our main study, we trained and deployed reward models that steer AI outputs \added{to minimise distortion of} writer stance, in order to test whether undesirable distortions can be mitigated without diminishing user preference for AI writing assistance.
Finally, with new panels of readers (N=8,798), we tested whether distortions carry downstream consequences, measuring how much readers trust AI-assisted writers and how readily they are persuaded by AI-assisted text depending on the degree of AI distortion.
Together, these studies offer the largest and most systematic assessment to date of how AI writing assistance distorts \added{perceptions of AI users and their opinions}, and provide an empirical foundation for understanding, measuring, and mitigating this phenomenon at scale. 

\begin{figure}[h]
    \centering
    \includegraphics[width=\linewidth]{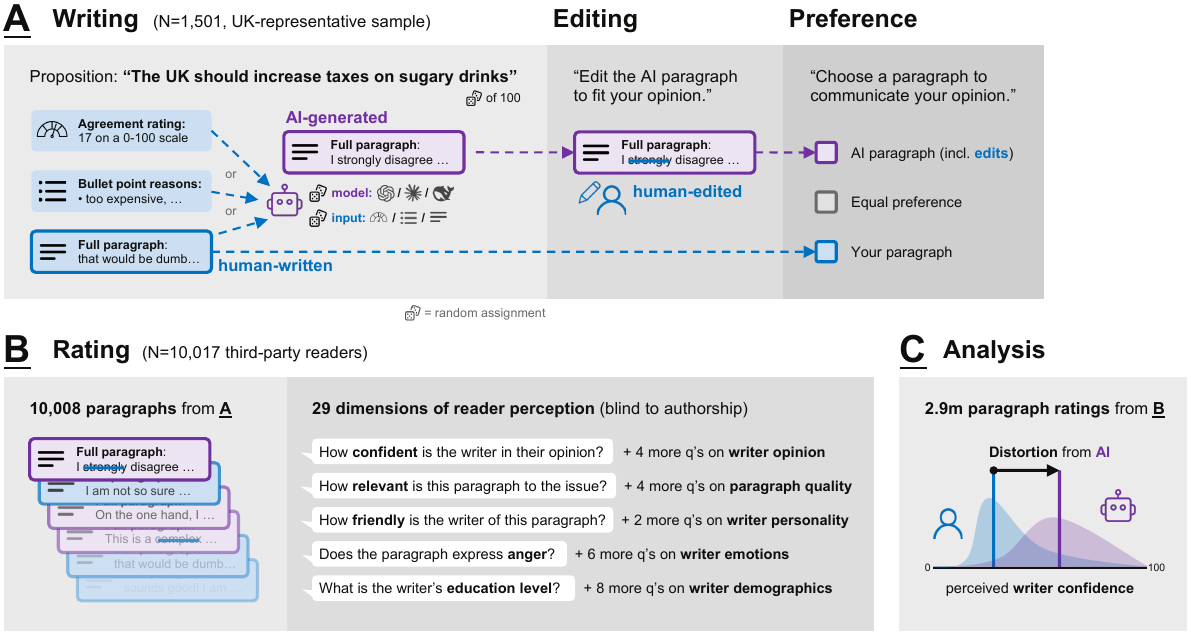}
    \caption{
    \textbf{Measuring distortions from AI writing assistance}.
    \textbf{\underline{A}:}~Writers expressed their opinion on a political proposition by rating their agreement on a 0-100 scale, listing reasons as bullet points, and writing an opinion paragraph.
    An AI model (randomly assigned) then generated another opinion paragraph based on one of these responses from the writer (also randomly assigned).
    Next, writers were asked to edit the AI paragraph until it reflected their opinion.
    Finally, writers chose whether they preferred the paragraph they had written or the edited, AI-generated paragraph for communicating their opinion.
    \textbf{\underline{B}:}~A separate sample of blinded readers evaluated all paragraphs across 29 dimensions of reader perception.
    \textbf{\underline{C}:}~Distortions from AI writing assistance are measured as systematic differences in reader perception between human-written and AI-generated paragraphs (with human edits) from the same writers.
    Sample sizes shown are for the main study (Study 1):\ 1,501~writers, 10,017~readers, 10,008~paragraphs, 2,903,596~ratings.
    }
    \label{fig:1}
\end{figure}

\section*{Results}
\label{sec:results}

In our main study (Study 1), writers were UK adults (N=1,501, census-representative on age, gender, race) who expressed their opinion on 3 political propositions drawn randomly from a pool of 100 (see \hyperref[sec:methods]{Methods}).
The propositions covered mainstream UK political issues balanced across the political spectrum, from healthcare and immigration to climate policy and civil liberties (see \textcolor{darkblue}{SM:\ref*{sm-subsec:study_materials_propositions}} for the full list).
Writers first rated their agreement with each proposition on a 0-100 scale, outlined their reasoning in two or more bullet points, and then expanded the bullet points into a full opinion paragraph of at least 100 words.
For each proposition assigned to the writer, one of these three writer inputs (rating, bullets, or paragraph) was passed to one of three AI models (Claude, DeepSeek, or ChatGPT) which generated a paragraph matching the format of the writer's own (see \hyperref[sec:methods]{Methods}).
To mirror everyday use of writing assistants, we then asked writers to edit the AI-generated paragraph until it reflected their opinion.
Writers edited the AI-generated paragraphs only 23\% of the time (<30\% across all models and input types at p<.01; see \textcolor{darkblue}{SM:\ref*{sm-subsec:main_preference_analysis_descriptive}}), and most edits were minor (median Levenshtein ratio = 0.96).
Writers reported moderate-to-high engagement with their assigned propositions (median issue knowledge = 56.0, median issue importance = 65.0, median confidence = 74.0 on 0-100 scales; see \textcolor{darkblue}{SM:\ref*{sm-sec:main_engagement_analysis}}), suggesting that our random assignment of propositions did not systematically force writers to opine on issues they felt uninformed about or indifferent toward.
\added{Our main results below are robust to excluding writer-proposition pairs for which writers reported low (<50) issue knowledge, importance, or confidence (see \textcolor{darkblue}{SM:\ref*{sm-subsec:main_distortion_analysis_engagement_robustness}})}.

\subsection*{Writer preference for AI writing}
\label{subsec:writer_preference}

Before addressing our main research questions, we tested a key empirical precondition for our work:\
that writers often accept and endorse AI-assisted writing as reflective of their own views.
If this was not the case, distortions introduced by AI would rarely propagate in the real world.

Thus, after writers had composed their own paragraph and edited the AI-generated version to their satisfaction, we asked them which version they preferred for communicating their opinion.
Writers strictly preferred the AI paragraph to their own in a clear majority of cases (2,835 of 4,503 cases, 63.0\%), and this preference held across AI models and writer inputs (preference rate >50\% across all conditions at p<.01; see \textcolor{darkblue}{SM:\ref*{sm-subsec:main_preference_analysis_descriptive}}).
Since writers had already composed their own opinion paragraph, this preference is unlikely to reflect mere convenience. 
When asked directly, in a majority of cases (1,477 of 2,835, 52.1\%) writers said they preferred AI writing because it better reflected their opinion than what they had written themselves.
This is consistent with prior work showing that writers \added{value AI writing assistance~\citep{li2024value} and} maintain a clear sense of control and agency over AI-assisted writing~\citep{kadoma2024role,bhat2026reactivewriters}.

\subsection*{Distortions from AI writing assistance}
\label{subsec:distortions}

Having established that writers routinely endorse AI-assisted text as reflective of their views and prefer it to their own writing, we turned to our first research question:\
does AI writing assistance systematically distort how others perceive writers and their opinions (\textbf{RQ1})?
To answer this question, we recruited a separate sample of readers (N=10,017 UK adults) who blindly rated both human-written and AI-assisted paragraphs across 29 dimensions of reader perception (see \hyperref[sec:methods]{Methods}). 
Distortions were then measured as systematic differences in reader perceptions between human-written paragraphs and AI-generated paragraphs (with human edits) from the same writers.
Importantly, in our main analysis (\hyperref[fig:2]{Figure~\ref*{fig:2}})
i)~the AI-generated paragraphs included all edits made by writers; and
ii)~cases where writers strictly dispreferred the AI-generated paragraphs (708 of 4,503 cases, 15.7\%) were excluded.
These criteria allowed us to measure distortions only where they would plausibly be propagated in the real world.
Results are robust to relaxing both restrictions (see \textcolor{darkblue}{SM:\ref*{sm-subsec:main_distortion_analysis_data_subsets}}).

AI writing assistance produced significant distortions across every dimension we measured (p<.01 each after Bonferroni correction across all 29 rating attributes; \hyperref[fig:2]{Figure~\ref*{fig:2}}).
With AI, writers seemed more extreme in their political opinions (+4.5 average marginal effect [AME] on a 0-100 scale), \added{slightly} less open to changing their views (-0.8), and \added{substantially} more confident (+7.8).
AI elevated perceived writing quality, with paragraphs judged as clearer (+9.5), more informative (+23.4), and more relevant (+8.7).
It compressed emotional expression into a narrower, more agreeable register:\
writers appeared friendlier (+4.5) and \added{much} more optimistic (+9.6), expressing more hope (+9.1) and excitement (+4.2) but less anger (-3.7), disgust (-3.5), and \added{slightly less} fear (-0.7).
It also shifted inferred writer demographics towards a more privileged profile:\
writers appeared more educated (×5.3 odds ratio), higher-income (×4.4), \added{much} more likely to be native English speakers (×4.1) and \added{slightly} more likely to be White (×1.1).
Effects across all 29 dimensions are shown in \hyperref[fig:2]{Figure~\ref*{fig:2}}.

AI writing assistance also homogenised perceptions of writers and their opinions.
Across most dimensions, AI-assisted paragraphs were rated significantly more similarly to each other than their human-written counterparts (signif.\ reduction in standard dev.\ [scale attributes] or entropy [categorical] for 22/29 rating attributes at p<.001 after Bonf.\ correction; see \textcolor{darkblue}{SM:\ref*{sm-subsec:main_distortion_analysis_homogenisation}}).
\added{This homogenising effect occurred in part because AI distortion was stronger when human writing was weaker (neg.\ interaction between AI use and human paragraph perception for all 20 scale and 5 ordinal attributes at p<.01 after Bonf.\ correction; see \textcolor{darkblue}{SM:\ref*{sm-subsec:main_distortion_analysis_baseline}}).
For example, writers whose own paragraphs made them seem relatively unconfident (10th \%ile) seemed much more confident with AI assistance (+18.3 AME), whereas reader perception of writers who already seemed confident (90th \%ile) changed little (+2.6).}
Overall, perceived writer confidence \added{thus} converged toward a narrower, more confident range for AI-assisted writing (24.1 vs 20.5 SD\added{, p<.001}).
This \added{finding on perceptual homogenisation} extends prior evidence of homogenisation at the lexical and semantic level~\citep{anderson2024homogenization,padmakumar2024does,sourati2026homogenizing}.

Notably, these effects were consistent across AI models and levels of human input.
On nearly every measure, all three models in our study (Claude, DeepSeek, ChatGPT) produced significant distortions in the same direction (18/20 scale attributes and 5/5 ordinal attributes at p<.05; see \textcolor{darkblue}{SM:\ref*{sm-subsec:main_distortion_analysis_effects_by_model}}), and in no case did two models produce significant distortions in opposite directions.
While exact distortion magnitudes varied across models (significant difference for at least one model pair for 18/20 scale attributes and 4/5 ordinal attributes at p<.05, \added{Holm-adjusted}; see \textcolor{darkblue}{SM:\ref*{sm-subsec:main_distortion_analysis_effects_by_model}}), no single model was consistently most distortive (avg.\ abs.\ AME = 8.8, 8.7, 8.6 for Claude, DeepSeek, ChatGPT across the 20 scale attributes; see \textcolor{darkblue}{SM:\ref*{sm-subsec:main_distortion_analysis_effects_by_model}}), suggesting that the patterns we observe reflect properties of current AI writing assistance in general rather than the design choices of any one model provider.
Similarly, distortions persisted across different levels of human input. 
They were strongest when models received only a numeric stance rating (avg.\ abs.\ AME of 10.5 across the 20 scale attributes; see \textcolor{darkblue}{SM:\ref*{sm-subsec:main_distortion_analysis_effects_by_input}}) or bullet points (10.2), but clearly present even when models received the writer's full paragraph (6.0 when model was asked to ``improve'' paragraph, 4.8 when asked to ``rewrite'').
That distortions persist even for full-paragraph input aligns with recent evidence of AI models altering textual meaning even when explicitly instructed to make only grammatical corrections~\citep{abdulhai2026llms}.

\begin{figure}[!b]
    \centering
    \includegraphics[width=\linewidth]{main_figures/fig2.pdf}
    \caption{
    \textbf{Distortions from AI writing assistance}, measured as systematic differences in reader perception (N=10,017 readers) of AI-assisted writing vs human writing (N=10,008 paragraphs) from the same authors (N=1,501 writers) across 29 rating attributes in our main study (Study 1).
    \textbf{Panels A-C}:
    Violin plots show rating distributions for human-written and AI-assisted paragraphs, with vertical lines for per-group means.
    Pts are average marginal effects from beta regressions, on the original 0-100 rating scale (see \hyperref[sec:methods]{Methods}).
    \textbf{Panel D}:
    Heatmaps show rating distributions for human-written and AI-assisted paragraphs, with icons indicating per-group modal categories.
    Odds ratios are from ordinal logistic regression for ordinal attributes (e.g.\ writer age, income) and one-vs-all logistic regression for nominal attributes (e.g.\ writer gender, race).
    *** indicates significance at p<.001 after Bonferroni correction across all 29 rating attributes.
    }
    \label{fig:2}
\end{figure}

\clearpage
Taken together, our results suggest that AI writing assistance introduces pervasive and consistent distortions, changing how writers \added{and their opinions} are perceived across a wide range of socially salient dimensions.
Writing with AI made writers seem more opinionated and more skilled, it compressed their emotional expression into a narrower and more agreeable register and shifted their perceived demographic profile towards more privileged groups.
It also homogenized reader perceptions of writers and their opinions.
But \added{how do writers feel about these effects}?

\subsection*{Writer tolerance for distortions from AI}
\label{subsec:tolerance}

Not all distortions from AI need be unwelcome.
Writers might accept, or even value, AI assistance that makes them appear more knowledgeable or confident.
Whether distortions are objectionable to the writers who propagate them has direct implications for writer agency and the wider normative concerns that distortions warrant (\textbf{RQ2}).
We thus asked each writer in our main study (N=1,501) how acceptable or unacceptable they would find different kinds of distortion, from AI making their writing clearer to AI making their political stances seem more extreme (see \hyperref[sec:methods]{Methods}).

\begin{figure}[h]
    \centering
    \includegraphics[width=\linewidth]{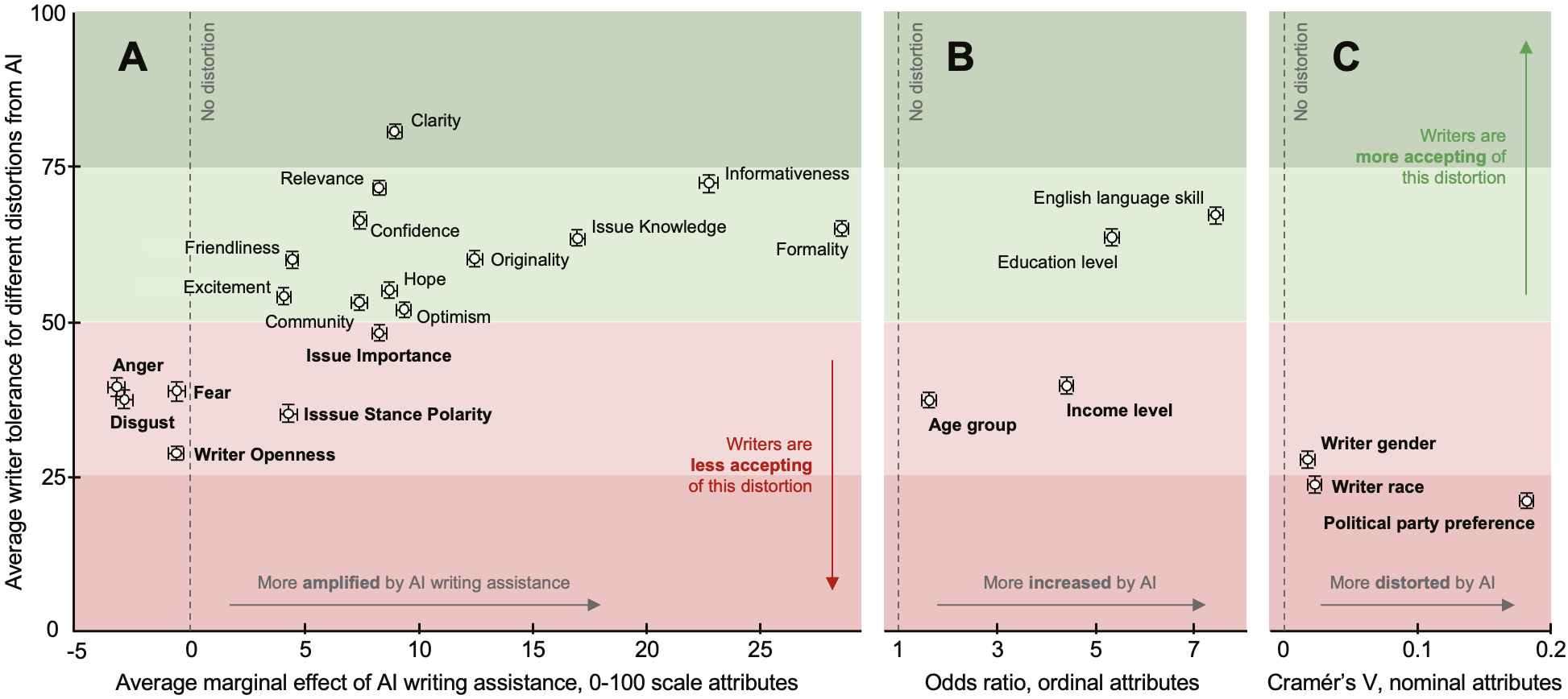}
    \caption{
    \textbf{Variation in writer tolerance for observed distortions from AI writing assistance}.
    Y-axis shows average stated tolerance of writers in our main study (Study 1, N=1,501) for different types of distortions, with 95\% bootstrap CIs.
    X-axes show magnitude of distortion from AI, with metrics differing by type of rating attribute.
    \textbf{A} for rating attributes measured on a 0-100 scale:\ Average marginal effect (AME) from mixed-effect beta regressions, with 95\% delta-method CIs.
    AME is positively correlated with writer tolerance (Pearson's r = 0.68, p=.002).
    \textbf{B} for ordinal rating attributes: Odds ratios from mixed-effect ordinal logistic regressions, with 95\% Wald CIs.
    \textbf{C} for nominal rating attributes: Cramer's V from comparing rating distributions for human-written and AI-assisted paragraphs, with 95\% bootstrap CIs (see \hyperref[sec:methods]{Methods})
    }
    \label{fig:3}
\end{figure}

Some distortions from AI writing assistance were welcome to writers (\hyperref[fig:3]{Figure~\ref*{fig:3}}).
Writers were most accepting of AI improving the perceived quality of their writing.
They felt positive about appearing clearer (80.8 avg.\ tolerance), more informative (72.5), and more relevant (71.7), and AI had particularly strong effects across these dimensions (+9.5, +23.4, +8.7 AME).
More broadly, the distortions that writers found most acceptable tended to be the largest in magnitude (\hyperref[fig:3]{Figure~\ref*{fig:3}A}).
Other distortions, however, were unwelcome.
Writers generally disliked the idea of AI distorting their perceived political opinion, emotions, or demographics, yet these are exactly the kinds of distortion we observed.
Most writers, for example, did not want AI to make their political stances seem more extreme (35.0 avg. tolerance), but we found clear evidence of this polarising effect (+4.3 AME).

These findings point to a tension between stated preferences and observed behaviour.
Writers objected to many of the distortions introduced by AI, yet they routinely preferred AI writing to their own and endorsed it as reflective of their views.
In doing so, they allowed AI to distort how they were perceived, including in ways they themselves described as objectionable.

\subsection*{Writer awareness of distortions}
\label{subsec:disclaimer}

Having observed that writers endorse AI-assisted text despite objecting to many of its distortions, we considered two possible explanations.
The first is ignorance:\ writers may simply not notice unwelcome distortions, or may not be aware that such distortions can occur.
The second is an implicit trade-off:\ writers may recognise that AI distorts their writing but accept this as a cost of perceived benefits such as improved clarity.
To distinguish between these accounts, we conducted a first follow-up study.

Specifically, for Study 2, we recruited a new sample of writers (N=669) who followed the same protocol as in Study~1, with one addition:\ a pop-up disclaimer (analogous to the generic accuracy warnings now standard in major AI products such as ChatGPT~\citep{henestrosa2025always}) that writers had to acknowledge before editing the AI paragraph and again before choosing their preferred paragraph version.
The disclaimer was randomly assigned at the writer level to highlight distortions that writers in Study~1 were least accepting of (e.g.\ ``AI tends to make your political opinion seem more extreme''), most accepting of (e.g.\ ``AI tends to make your writing clearer''), or both (see \hyperref[sec:methods]{Methods}).

We found that these disclaimers had no reliable impact on the tendency of writers to edit the AI-written paragraphs (23.1\% edit rate in no-disclaimer control, no disclaimer condition significant, p>.05; see \textcolor{darkblue}{SM:\ref*{sm-subsec:disclaimer_study_stats}}) nor did they change the rate at which writers strictly preferred AI writing to their own (57.8\% in no-disclaimer control, no disclaimer condition significant, p>.05; see \textcolor{darkblue}{SM:\ref*{sm-subsec:disclaimer_study_stats}}).
Ignorance of the potential for distortions therefore seems unlikely to explain why writers prefer AI writing.
Even when explicitly warned that AI may distort their writing in ways they were likely to object to, writers continued to endorse AI outputs as reflective of their views.

\subsection*{Targeted mitigation of distortions}
\label{subsec:mitigation}

Study~2 showed that informing writers about distortions does not change their preference for AI-assisted text, which suggests that user-facing interventions cannot curb the propagation of unwelcome distortions.
Thus, we ask:\ can model-level interventions reduce distortions at the source (\textbf{RQ3})?
To answer this question, we conducted a second follow-up study (Study 3).

Here, we targeted the polarising effect of AI writing assistance — AI making the writer's stance seem more politically extreme — because it was both pronounced and particularly objectionable to writers.
We recruited a new sample of writers (N=769), who followed the same experimental protocol as in Studies 1-2, and a new sample of readers (N=2,543) who rated all paragraphs across 29 attributes.

As a first intervention, we tried Prompting:\ appending a short instruction (113 words; see \textcolor{darkblue}{SM:\ref*{sm-subsec:mitigation_study_prompting_design}}) to the AI model's generation prompt, explicitly directing it to preserve the writer's issue stance.
This is a minimal intervention but a plausible one, since AI models are explicitly optimised to follow natural language instructions~\citep{ouyang2022training} and can, in principle, self-correct undesirable tendencies when directed to do so~\citep{schick2021self}.
However, Prompting was ineffective.
Compared to a No Intervention control, it did not significantly reduce the polarising effect of AI writing assistance (+7.4 vs 6.2 AME, p=.08).

\begin{figure}[htb]
    \centering
    \includegraphics[width=0.98\linewidth]{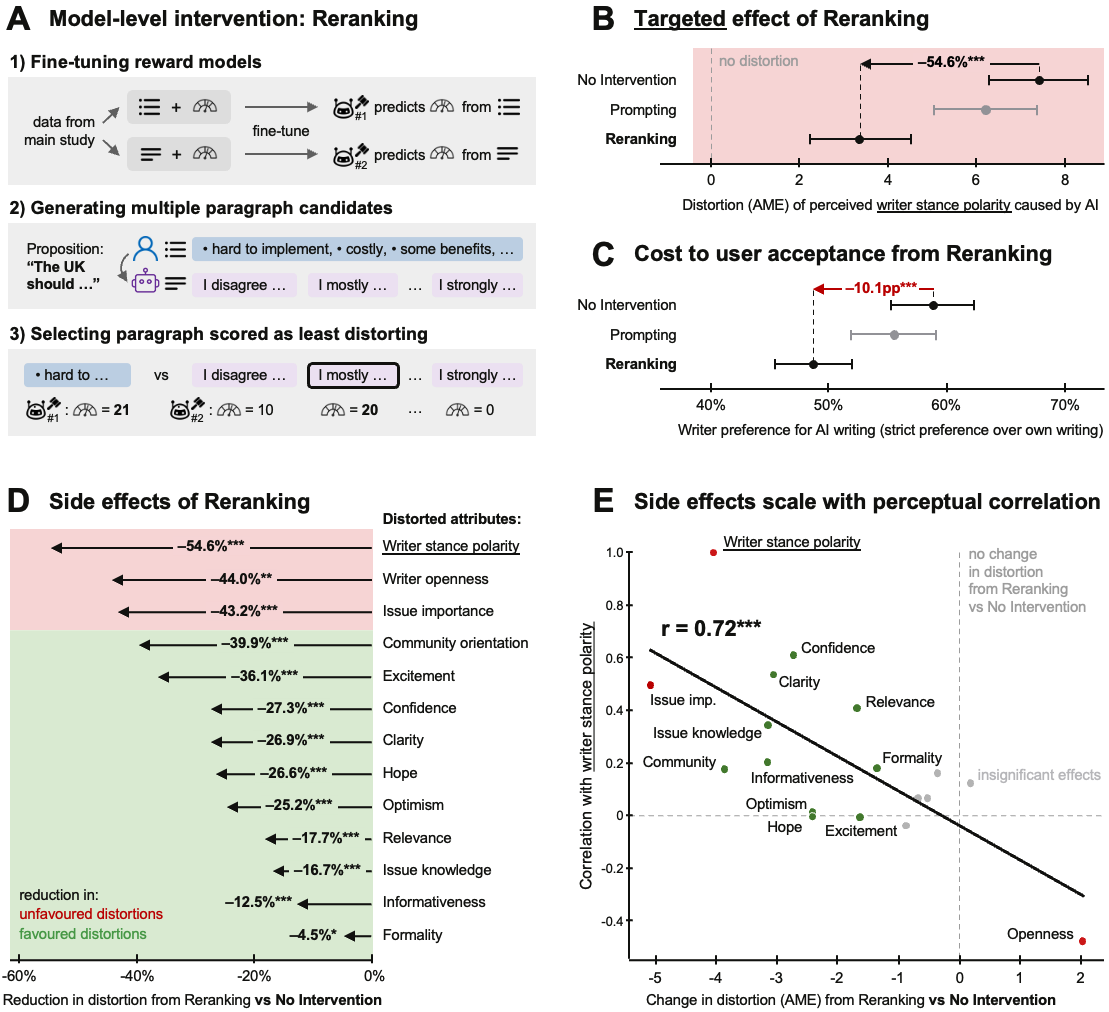}
    \caption{
    \textbf{Mitigating distortion from AI writing assistance}.
    \textbf{A}:\ Design of the Reranking intervention.
    1)~We fine-tune two reward models on reader ratings from Study~1 to predict perceived issue stance for human-written and AI-assisted paragraphs.
    2)~At generation time, the AI model generates multiple paragraph candidates.
    3)~We select the AI candidate predicted to minimise stance distortion relative to the writer's own paragraph.
    \textbf{B}:~Reranking succeeded in reducing the targeted polarising distortion while Prompting failed to do so (p=.08).
    Error bars are 95\% CIs from beta regression.
    \textbf{C}:~ Reranking reduced the rate at which writers strictly prefer AI-assisted paragraphs to their own.
    Error bars are 95\% bootstrap CIs.
    \textbf{D}:~While Reranking succeeded in reducing distortion in writer stance polarity, it also attenuated a range of non-targeted distortions that writers in Study~1 found acceptable (green) and unacceptable (red).
    \textbf{E}:~Side effects of Reranking scaled with how correlated reader perceptions on those non-targeted attributes were with perception of stance polarity, the target attribute.
    The diagonal line shows a linear fit of the plotted points, with Pearson's r = 0.72.
    * p<.05, ** p<.01, *** p<.001.
    }
    \label{fig:4}
\end{figure}

Given that neither user-facing disclaimers nor direct prompting mitigated the polarising distortion, we turned to a more sophisticated model-level intervention:\ Reranking (\hyperref[fig:4]{Figure~\ref*{fig:4}A}).
This approach draws on best-of-N selection methods used in AI training, where reward models steer outputs toward a specified objective by scoring and selecting among multiple output candidates~\citep{bai2022constitutional,ouyang2022training}.
Similar reranking strategies have recently been applied to reduce or exacerbate properties of AI-generated text in other domains, including the political persuasiveness of conversational AI~\citep{hackenburg2025levers}.
We adapted this logic to \added{reduce distortion} of writer stance.
Specifically, we fine-tuned two reward models (RMs) using the 10,008 annotated paragraph ratings collected in Study~1:\ one to predict perceived issue stance from AI-assisted paragraphs, and another to predict the perceived stance of the writer's own text from their bullet point inputs.
At generation time, we used verbalised sampling~\citep{zhang2026verbalized} to generate multiple AI paragraph candidates, and selected the candidate predicted by our RMs to minimise stance distortion relative to the writer's own paragraph (see \hyperref[sec:methods]{Methods} for details).
Reranking successfully reduced the targeted polarising effect.
Compared to No Intervention, it shrank the polarising distortion by 54.6\% (+7.4 vs +3.4 AME, p<0.01; \hyperref[fig:4]{Figure~\ref*{fig:4}}B), demonstrating that distortions from AI writing assistance can be substantially reduced at the model level.

However, this targeted reduction came at a cost.
Writers were significantly less likely to prefer AI-assisted paragraphs produced under Reranking than under No Intervention (48.8\% vs 58.9\% strict preference, p<.001; \hyperref[fig:4]{Figure~\ref*{fig:4}}C).
This drop in preference is likely explained by Reranking's side effects on non-targeted dimensions of reader perception (\hyperref[fig:4]{Figure~\ref*{fig:4}}D).
In addition to reducing the unwelcome polarising distortion, Reranking attenuated several distortions that writers found acceptable, such as appearing more confident, clearer, and more knowledgeable (-27.3\%, -26.9\%, and -16.7\% reduction in distortion from Reranking vs No Intervention, all p<.001).

This pattern of entangled distortions suggests a deeper structure in how readers form impressions from text.
Across all our studies, reader perceptions on different dimensions were strongly intercorrelated. 
Perceived writer polarity, for instance, was associated with how confident, clear, and knowledgeable writers appeared (Pearson's r = 0.60, 0.52, 0.28, all p<.001; full correlation analyses in \textcolor{darkblue}{SM:\ref*{sm-subsec:main_distortion_analysis_correlations}}).
When we intervened to reduce the polarising effect, distortions on correlated dimensions were attenuated in proportion:\ the degree of attenuation for each non-targeted distortion was predicted by its correlation with stance polarity (\hyperref[fig:4]{Figure~\ref*{fig:4}}E; Pearson's r = 0.72, p<.001).
Such entanglement between desirable and undesirable properties of model outputs echoes trade-offs observed in other AI alignment contexts, where optimising for one objective can degrade performance on correlated objectives~\citep{bai2022constitutional,gao2023scaling}.

Overall, these results suggest that the side effects of Reranking are not arbitrary but arise from a shared perceptual structure underlying reader judgments.
Writers may object to the polarising effect in isolation, but they also value distortions that are, to a considerable extent, perceptual correlates of the same underlying shift.
Model-level interventions that reduce undesirable distortions from AI writing assistance therefore risk attenuating desirable ones, creating a trade-off between distortion reduction and user acceptance that may be difficult to fully resolve.

\subsection*{Downstream consequences of distortions}
\label{subsec:trust/persuasion}

So far, we have shown that perceptions of writers and their opinions are systematically distorted by AI writing assistance.
Distortions from AI are pervasive and difficult to mitigate.
But do these perceptual shifts affect human behaviours and attitudes (\textbf{RQ4})?
To answer this question, and thus validate the practical relevance of our earlier findings, we conducted two more follow-up studies.

In Study~4, we recruited another new sample of readers (N=802) and measured how much trust they placed in AI-assisted vs.\ unassisted writers.
Specifically, each reader played four rounds of a Trust Game~\citep{berg1995trust}, in which they decided how much of a real 20p endowment to allocate to the author of an assigned paragraph. 
This author, readers were told, would receive 3$\times$ what they allocated, and could then decide how much to return.
Readers would be paid whatever portion of the endowment they kept plus whatever the author returned. 
Larger allocations from the reader thus indicate higher trust in the paragraph author.
The paragraphs were drawn from Study~1 and balanced across two factors:\ who wrote the paragraph (human vs.\ AI), and how much distortion the AI introduced relative to the writer's own paragraph (low vs.\ high, based on Study 1 ratings averaged across all 20 scale attributes).
Each reader saw one paragraph from each of the four resulting cells (see \hyperref[sec:methods]{Methods}).

In Study~5, we again recruited new readers (N=7,996) and measured how persuaded they were in their stance on political propositions by human- vs.\ AI-written paragraphs.
The paragraphs shown to readers, and the corresponding political propositions, were sampled from the same set as Study~4, plus a fifth, off-topic control paragraph with no political stance (see \hyperref[sec:methods]{Methods}).
Each reader rated their agreement (0-100 scale) before and after reading an assigned paragraph for five different propositions.

\begin{figure}[h]
    \centering
    \includegraphics[width=\linewidth]{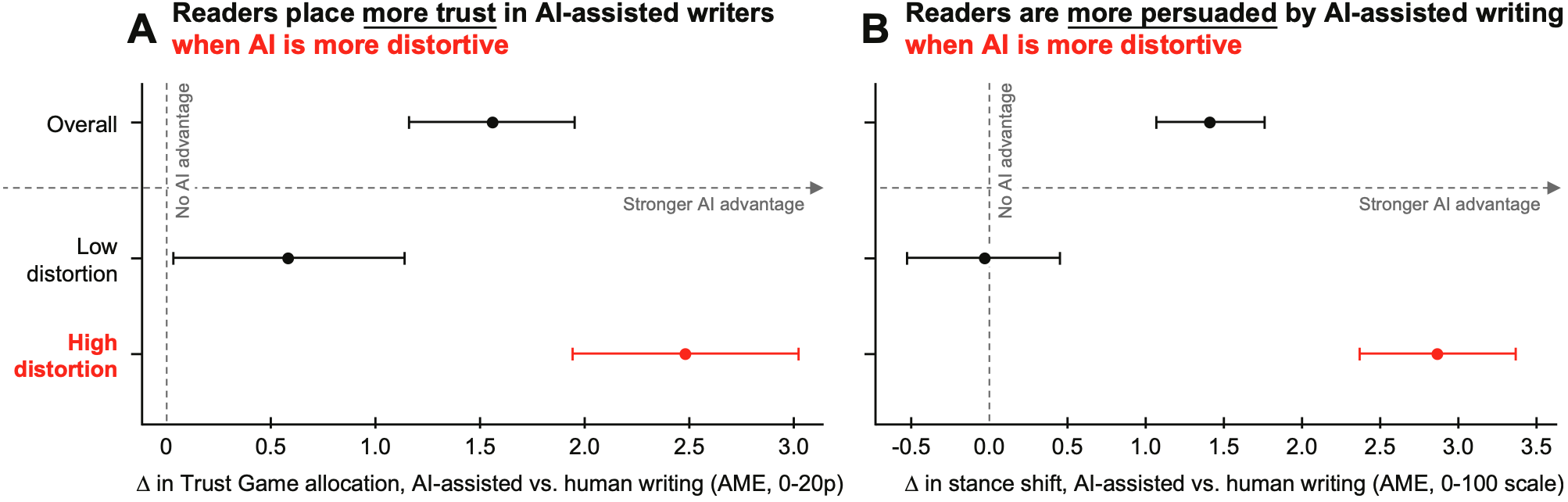}
    \caption{
    \textbf{Effects of \added{AI-assisted writing} on reader behaviours and attitudes}, overall and across different amounts of distortion from AI.
    \textbf{A} (Study~4, N=802) for reader trust in the paragraph author as measured by Trust Game allocations (0-20p):\ average marginal effect (AME) for AI-assisted vs.\ human writing from mixed-effect beta regressions, with 95\% Wald CIs.
    \textbf{B} (Study~5, N=7,996) for paragraph persuasiveness as measured by pre/post stance shift towards the paragraph stance (0-100 scale):\ AME for AI-assisted vs.\ human writing from mixed-effect linear regressions, with 95\% Wald CIs (see \hyperref[sec:methods]{Methods}).
    }
    \label{fig:5}
\end{figure}

We found that readers placed significantly more trust in AI-assisted writers than their unassisted counterparts (\hyperref[fig:5]{Figure~\ref*{fig:5}}A; overall AME of +1.56p in Trust Game allocation, p<.001).
AI-assisted writing also had a significantly stronger persuasive impact on readers than human writing (\hyperref[fig:5]{Figure~\ref*{fig:5}}B; overall AME of +1.41 in stance shift, p<.001), matching evidence from prior work on AI persuasion~\citep{goldstein2024persuasive,bai2025persuasion,hackenburg2025levers}.
Notably, both effects scaled with distortion magnitude.
When AI assistance introduced little distortion relative to human writing, there was little AI advantage in terms of trust (+0.58p, p=.039) and no significant advantage in persuasiveness (p=.88).
When AI was very distortive, however, it made writers appear significantly more trustworthy (+2.49p, p<.001) and was much more effective at persuading readers than human writing (+2.87, p<.001).
This result replicates for a continuous measure of distortion magnitude (+0.98p Trust Game allocation and +1.47 stance shift per 1 SD increase in distortion from AI writing assistance, both p<.001; see \textcolor{darkblue}{SM:\ref*{sm-subsec:trust_study_statistical_analysis}} and \textcolor{darkblue}{SM:\ref*{sm-subsec:persuasion_study_statistical_analysis}}).

\added{Together, these findings show that perceptual distortions from AI writing assistance coincide with consequential effects on reader behaviours and attitudes.
AI effects on trust and persuasion were strongest precisely where AI most changed perceptions of writers and their opinions, suggesting that these effects follow specifically from \textit{distortive} AI rather than from AI assistance as such.}

\section*{Discussion}
\label{sec:discussion}

We found that AI writing assistance \added{distorts perceptions of writers and their opinions} across a wide range of socially salient dimensions.
Writing with AI made writers seem more opinionated, more competent, more positive, and more likely to come from a privileged background.
Readers placed greater trust in AI-assisted writers and were more readily persuaded by \added{AI-assisted writing}, particularly when AI was more distortive.
With monthly requests for AI writing assistance now rising to the tens of billions~\citep{chatterji2025people}, distortions could thus have powerful effects on human interpersonal relationships and societal dynamics.

One particularly striking finding was that AI\added{-assisted} writing made human users appear more opinionated, i.e.\ more extreme and more confident in their views.
This is concerning given that AI is already used to write political speeches delivered in national legislatures~\citep{pimlico2025chatgpt}, news articles published in major media outlets~\citep{harrington2023cnet}, legal arguments presented in court~\citep{wu2025federal} as well as academic publications~\citep{he2026academic} and peer reviews~\citep{abdulhai2026llms}.
In an age where opinions on social, cultural and political issues are becoming more polarised, potentially fuelled by access to media that algorithmically prioritise more extreme views~\citep{milli2025engagement}, this distortion could further intensify political division, compounded by the increased persuasiveness of AI writing.
While our participants did not welcome the idea that AI writing assistance might \added{distort} their opinions, they accepted distorted AI-written text and passed up opportunities to edit it to more faithfully reflect their views, suggesting that human review offers at best a weak safeguard against a wave of more opinionated political writing~\citep{bhat2026reactivewriters}.

Other observed distortions, such as writers appearing more knowledgeable or educated when using AI assistance, might seem more innocuous.
\added{Indeed, prior work has framed positive, homogenising effects of AI use
as potentially equity-enhancing~\citep{noy2023experimental,wiles2025algorithmic,tukachinsky2026digital}.
However, in a world of unequal AI access and adoption~\citep{chatterji2025people,bick2026rapid,hartley2026labor}, advantages from AI writing assistance may exacerbate existing inequalities~\citep{humlum2025unequal}.
In aggregate, seemingly positive distortions may also contribute to an erosion of the signalling value of text regarding writer character and competence. 
Employers, for example, may already be mistrustful of written submissions made during hiring, shifting to alternative indicators instead~\citep{cui2025signaling,galdin2025making}.
Academic assessment and scientific review, similarly, are in crisis precisely because AI use increasingly decouples surface-level fluency from genuine understanding and scientific merit~\citep{messeri2024artificial,bastani2025generative,kusumegi2025scientific}.
}

Distortions from AI writing assistance most likely arise because modern AI models are optimised to produce outputs that are favoured by human raters~\citep{ouyang2022training}, who may prefer AI responses that seem more well-informed, confident, or upbeat.
Indeed, the clear preference for AI writing among our participants as well as the widespread adoption of AI writing assistance~\citep{chatterji2025people,handa2025economic,tomlinson2025working} suggest that this optimisation succeeds.
However, the distortions that writers favour appear closely entangled with other distortions that writers would rather avoid.
Writers liked seeming more confident, but disliked seeming more extreme. 
Unfortunately, this meant that mitigations meant to target dispreferred qualities of AI writing assistance (such as extremity) also reduced favoured qualities (such as confidence), which dampened overall enthusiasm for AI writing.
Unless future work can resolve this trade-off, AI developers will have little incentive to train \added{less distortive} AI writing assistants, despite the risks \added{they pose}.

Our study has several limitations that qualify the interpretation of these findings.
In real-world settings where personal or professional stakes are higher, writers may scrutinise AI-generated text more carefully and edit more extensively than participants in our study did.
Readers, likewise, may have access to additional cues about the writer, repeated interactions, and stronger incentives to form accurate impressions, all of which could attenuate or amplify the distortions we document.
We focused on political opinion writing, where identity and stance are particularly salient, but distortions may manifest differently in professional, creative, or interpersonal writing.
We recruited only UK participants, but \added{political writing as well as} the nature and magnitude of distortions from AI writing assistance may differ across languages and cultures~\citep{henrich2010most}.
\added{Readers' perceptions may also differ if they were aware that texts were AI-generated \citep{jakesch2019ai,altay2024people,schilke2025transparency} (although, absent explicit disclosure, readers generally struggle to distinguish AI- from human-written text~\citep{clark2021humanisnotgold,jakesch2023human,williams2025large}).}
Finally, future AI models may produce different patterns of distortion, as a consequence of new training methods, \added{increased model personalisation, or} shifting developer priorities.
However, the clear preference for distortive AI writing assistance among our participants, and the consistency in distortions we observed across the three leading AI models we tested, suggest that the broad phenomenon is likely to persist in the future.

\section*{Methods}
\label{sec:methods}

This research was approved by a Responsible Research Process at the UK AI Security Institute and Bocconi University's Ethics Committee (RA000863.01).
Overall, we ran five studies.
The main study (Study 1) was conducted between June 13th and July 21st, 2025;
the disclaimer study (Study 2) between November 25th and December 3rd, 2025;
the mitigation study (Study 3) between December 16th, 2025 and January 29th, 2026;
the trust study (Study 4) on June 1st and 2nd, 2026; and
the persuasion study (Study 5) between June 4th and 11th, 2026.
All studies were preregistered (1:\ \href{https://osf.io/6exf2/overview}{osf.io/6exf2}, 2 \& 3:\ \href{https://osf.io/sf7rg/overview}{osf.io/sf7rg}, 4: \href{https://osf.io/38sg7/overview}{osf.io/38sg7}, 5: \href{https://osf.io/e48g2/overview}{osf.io/e48g2}).
We note where analyses deviate from pre-registration.
All code and replication materials are available in our \href{https://github.com/paul-rottger/ai-distortion}{project repository}.

\subsection*{Participants}

All participants were adults (18+ years old) based in the UK, recruited via the online crowd-sourcing platform Prolific.
We distinguish participants between writers, who shared their political opinions and engaged with AI-assisted versions of their writing, and readers, who rated or otherwise engaged with human- and AI-written opinion paragraphs.
For writers, we targeted UK census-representative samples based on age, gender, and race.
For readers, we did not require representativeness due to the large size of our target sample.
Study~1 was completed by 1,501 writers and 10,017 readers, Study~2 by 669 writers, Study~3 by 769 writers and 2,543 readers, Study~4 by 802 readers, and Study~5 by 7,996 readers.
There was no overlap between writers and readers (i.e.\ no writer was also a reader, and vice versa).
Each participant took part in only one study, with one exception made to reach target sample size:\
1,469 readers in Study~3 had previously participated as readers in Study~1 approximately five months earlier.
Thus, overall, there were 2,939 distinct writers and 19,889 distinct readers.
For participant demographics and UK census comparisons, see \textcolor{darkblue}{SM:\ref*{sm-subsec:participants}}.
For attrition analysis, see \textcolor{darkblue}{SM:\ref*{sm-subsec:attrition}}.

\subsection*{Political propositions}

In Studies 1-3, writers expressed their opinion on 3 political propositions, randomly selected from a pool of 100.
We constructed these propositions based on political issues extracted from real user conversations with AI models in prior work~\citep{rottger2026issuebench}.
We selected political issues and adapted propositions to be relevant to a broad UK audience, specific, and clear in stance (e.g.\ ``The UK should mandate COVID-19 vaccinations for all eligible citizens.'').
By design, the overall set of propositions is balanced in political leaning, in that there is an equal number of propositions that clearly align with UK left- or right-wing views.
For details on proposition design and the full list of propositions, see \textcolor{darkblue}{SM:\ref*{sm-subsec:study_materials_propositions}}.

\subsection*{AI models}

In Studies 1-3, writers engaged with three AI models:\
Claude Sonnet 4~\citep{anthropic2025claude}, DeepSeek V3~\citep{liu2024deepseek}, and ChatGPT-4o~\citep{hurst2024gpt}, which is the version of GPT-4o used in ChatGPT.
We selected these models to represent the state-of-the-art in widely available AI chat assistants when we launched Study~1 in June 2025.
All inference was done via OpenRouter (endpoints:\ \texttt{anthropic/claude-sonnet-4}, \texttt{deepseek/deepseek-chat-v3-0324}, and \texttt{openai/chatgpt-4o-latest}).
Generation parameters were set to default values, including a temperature of 1, to match natural user interactions.

\subsection*{Writer inputs to AI models}

In Studies 1-3, writers expressed their opinion on each assigned political proposition in three formats:
i)~a numeric stance rating indicating agreement with the proposition on a 0-100 scale, where 0 = ``strongly disagree'', 50 = ``ambivalent'', and 100 = ``strongly agree''; 
ii)~bullet points ($\geq$2) explaining ``key reasons'' for their opinion; and
iii)~a full opinion paragraph ($\geq$100 words).
For each writer-proposition pair, one AI model was instructed to generate one opinion paragraph based on one of these three writer inputs:\
for stance and bullet inputs, to ``write a short paragraph of 100-150 words'', and for paragraph inputs to ``rewrite'' or ``improve'' the writer's paragraph (separate conditions, randomly assigned) ``without changing its length''.
\added{We varied writer inputs in this way to reflect real-world use of AI writing assistance, which ranges from generating content from minimal user input to rewriting or polishing near-complete drafts~\citep{longpre2026ai}.}
For the full generation prompts, see \textcolor{darkblue}{SM:\ref*{sm-subsec:study_materials_input_conditions}}.

In Studies 1 and 2, we randomly paired the three AI models with the three input formats, so that each writer engaged with content generated by each model and based on each input type, in a within-subjects design.
In Study~3, we used only bullet point inputs, as pre-registered, because we had found the targeted polarising effect of AI to be most pronounced under this condition in Study~1.

\subsection*{Writer preference analysis}

In Studies 1-3, writers were asked to ``edit the AI-written paragraph until you feel that it reflects your opinion about the [proposition]''.
Then, writers were asked to choose between their own writing and the AI-written paragraph, including any edits made, for ``explaining your opinion on the [proposition] to someone''.
Writers could choose strict preference for either paragraph or express equal preference.
\added{In cases of strict preference, they had to explain their reasoning either by selecting categorical options (Study~1) or in free text (Studies~2,~3; see \textcolor{darkblue}{SM:\ref*{sm-subsec:main_preference_analysis_reasons}}, \textcolor{darkblue}{SM:\ref*{sm-subsec:disclaimer_study_descriptive}}, \textcolor{darkblue}{SM:\ref*{sm-subsec:mitigation_study_writer_preference} for additional analysis).}}

From \added{the edit and preference} questions, we derived three binary variables observed for each writer-proposition pair:\
whether the writer made any edits, whether the writer weakly preferred AI writing, and whether the writer strictly preferred AI writing.
In addition to reporting edit and preference rates for Study~1, we fit logistic regression models with writer random effects to each binary outcome to measure how writer preference for AI writing varied across i)~which AI model was used and ii)~which type of writer input was passed to the AI.
For full specifications and regression outputs, see \textcolor{darkblue}{SM:\ref*{sm-subsec:main_preference_analysis_stats}}.
Note that our Study~1 analysis deviates from our pre-registration in using a binary edit indicator instead of continuous edit distance because edits were much rarer and smaller than we had anticipated.
We report results for the continuous edit distance outcome in \textcolor{darkblue}{SM:\ref*{sm-subsec:main_preference_analysis_stats}}.

\subsection*{Reader perception ratings}

In Studies 1 and 3, i.e.\ the main and mitigation studies, readers were each randomly assigned ten paragraphs, which they rated across 29 dimensions of reader perception that spanned political opinion, writing quality, writer personality, emotions, and demographics (see \textcolor{darkblue}{SM:\ref*{sm-subsec:study_materials_main_p2}} for exact question wordings).
\added{We selected these dimensions to capture distinct and likely consequential aspects of reader perception concerning texts and their writers.
Perceived writer confidence, for example, is known to affect persuasion, moderated by perceived expertise~\citep{karmarkar2010believe}.
Topical relevance is a core dimension in established taxonomies of argument quality~\citep{wachsmuth2017computational} (see \textcolor{darkblue}{SM:\ref*{sm-subsec:study_materials_rating_rationale}} for rationales on each of the 29 dimensions).}
In cases where writers made edits to the AI paragraphs, we included both the edited and unedited versions in the paragraph pool, which thus comprised 10,008 paragraphs in Study~1 and another 5,016 paragraphs in Study~3.
Readers were not told who authored the paragraphs, and there was no mention of possible AI involvement.
Assignment was such that, on average, each paragraph was rated by ten readers in Study~1 (\textcolor{darkblue}{SM:\ref*{sm-subsec:main_distortion_analysis_reader_assignment}}) and by five readers in Study~3 (\textcolor{darkblue}{SM:\ref*{sm-subsec:mitigation_study_reader_assignment}}).
We did not collect third-party ratings for Study~2, i.e.\ the writer-focused disclaimer study.

\subsection*{Distortion analysis}

In Studies 1 and 3, we assessed whether AI writing assistance caused a significant change in third-party perceptions of writers and their opinions.
For this purpose, we fit generalised linear mixed-effects regression models for each of the 29 rating attributes, where the single regressor was binary paragraph type (human-written or AI-assisted).
For the 20 rating attributes measured on a 0-100 scale (e.g.\ paragraph relevance, writer confidence) we fit beta regressions using the logit link function, after mapping each outcome to the (0,1) interval.
For the 5 ordinal rating attributes (e.g.\ writer income, education) we fit ordinal logistic regressions using the cumulative logit link function.
For the 4 nominal rating attributes (e.g.\ writer race, gender) we fit multinomial logistic regressions using the multinomial logit link function, \added{as well as one-vs-all logistic regressions to report category-specific effects.}
In all cases we included reader random effects \added{to account for the highly subjective nature of reader perceptions (see \textcolor{darkblue}{SM:\ref*{sm-subsec:main_distortion_analysis_rater_agreement}} and \textcolor{darkblue}{SM:\ref*{sm-subsec:mitigation_study_rater_agreement}} for inter-rater agreement analysis).}
\added{For the scale attributes, we additionally included writer and proposition random effects, although these explained less variance and left estimates essentially unchanged (\textcolor{darkblue}{SM:\ref*{sm-subsec:main_distortion_analysis_random_effects}})}
All hypothesis tests were two-sided.
For full specifications and regression outputs, see \textcolor{darkblue}{SM:\ref*{sm-subsec:main_distortion_analysis_overall_effects}}.

We quantified overall distortion using different measures derived from our fitted regression models.
For the 0-100 scale rating attributes, we used average marginal effects (AME) based on predictions from our beta regressions.
For the nominal and ordinal rating attributes, we used odds ratios as exponentiated coefficients from our ordinal and one-vs-all logistic regressions.

To measure variation in distortion across i)~which AI model was used and ii)~which type of writer input was passed to the AI in Study~1, we fit expanded versions of each regression that included AI and input condition regressors.
For full specifications and outputs, see \textcolor{darkblue}{SM:\ref*{sm-subsec:main_distortion_analysis_effects_by_model}} and \textcolor{darkblue}{SM:\ref*{sm-subsec:main_distortion_analysis_effects_by_input}}.

Note that our Study~1 regression analysis deviates from our pre-registration, which had specified a three-way comparison between all human-written paragraphs, all AI-written paragraphs excluding edits, and those AI-written paragraphs that were edited.
To simplify interpretation, we make two-way comparisons directly between human-written and AI-assisted paragraphs, where the latter include any edits made.
Results are robust to excluding edits, as shown in  \textcolor{darkblue}{SM:\ref*{sm-subsec:main_distortion_analysis_overall_effects}}.

\added{Results are also robust to features of our rating protocol.
Distortions were essentially unchanged when we controlled for the type of the previously rated paragraph, or excluded survey speeders and straightliners.
The direction of distortions also remained the same when we analysed only the first paragraph each reader rated, which is unaffected by potential fatigue or carryover effects (\textcolor{darkblue}{SM:\ref*{sm-subsec:main_distortion_analysis_rating_robustness}}).}

\added{In \textcolor{darkblue}{SM:\ref*{sm-subsec:main_distortion_analysis_perception_accuracy}}, we compared reader perceptions (e.g.\ on writer confidence) to writer self-reports on the same questions, finding that 
AI writing assistance made reader perceptions less accurate.}

\added{Finally, to complement our results on perceptual distortion, we ran textual analyses and replicated prior findings, e.g.\ that AI homogenises text at the lexical and semantic level, in \textcolor{darkblue}{SM:\ref*{sm-subsec:main_distortion_analysis_homogenisation}} and \textcolor{darkblue}{SM:\ref*{sm-subsec:main_distortion_analysis_text}}}.

\subsection*{Writer distortion tolerance analysis}

In our main study (Study~1, N=1,501 writers), we measured stated writer tolerance for different kinds of distortion from AI writing assistance.
We showed each writer a series of statements matching the rating attributes shown to readers (e.g.\ ``I am okay with AI writing assistance making my writing seem more relevant or on-topic.''), and then recorded writer agreement with each statement on a 0-100 scale, where 0 = ``strongly disagree'', 50 = ``ambivalent'', and 100 = ``strongly agree''.
For question wordings see \textcolor{darkblue}{SM:\ref*{sm-subsec:study_materials_main_p1}}, and for per-question rating distributions \textcolor{darkblue}{SM:\ref*{sm-sec:main_distortion_tolerance}}.

\subsection*{Disclaimer study}

In Study~2 (N=669 writers), we investigated whether making writers aware of potential distortions from AI writing assistance changes their preference for AI writing.
This study matched the Study~1 design, except for the addition of a pop-up disclaimer that writers had to acknowledge before they could edit the AI paragraph, and again before they chose which paragraph (AI or their own) they preferred for communicating their opinion.
The content of the disclaimer was randomly assigned at the writer level across three disclaimer conditions and a no-disclaimer control.

Disclaimer design was based on the writer tolerance measures collected in Study~1.
The first disclaimer (N=160 writers) highlighted the two observed AI distortions that, on average, writers were most accepting of:\ AI making writing seem clearer and more relevant to the issue that the paragraph was responding to.
The second disclaimer (N=154) highlighted the two observed AI distortions that, on average, writers were least accepting of:\ AI making writers seem more extreme in their political opinion and less open to changing their views.
The third disclaimer (N=185) highlighted all four of these observed AI distortions.
We chose this design to distinguish whether any effect of disclaimers on writer behaviour is driven by awareness of unwelcome distortions, welcome distortions, or distortions in general.
The order in which distortions were highlighted within each disclaimer was randomised every time a disclaimer was shown.
For exact disclaimer wording, see \textcolor{darkblue}{SM:\ref*{sm-subsec:disclaimer_study_design}}.

When analysing writer preference for AI writing in the disclaimer study, we used the same three binary outcome variables as in Study~1:\ whether writers made edits to the AI paragraph and whether they weakly/strictly preferred the AI paragraph to their own.
In addition to reporting edit and preference rates, we fit logistic regression models with a disclaimer condition regressor and writer random effects to each outcome variable.
For regression specifications and outputs, see \textcolor{darkblue}{SM:\ref*{sm-subsec:disclaimer_study_stats}}.

\subsection*{Mitigation study}

In Study~3 (N=769 writers), we piloted two model-level interventions to reduce distortions from AI writing assistance.
This mitigation study matched the Study~1 design, except for how paragraphs were generated by AI, where we assigned a model-level intervention (Prompting, Reranking) or a No Intervention control.
Assignment was such that each writer, across the three propositions they were assigned, experienced all three conditions in random order, in a within-subjects design.
We targeted the polarising effect of AI with our interventions (i.e.\ AI writing assistance making writers seem more extreme in their political opinion) because it was both pronounced and particularly undesirable to writers.
Our interventions are equally compatible, by design, with targeting other distortions, although their effectiveness may vary across distortion types.

For the Prompting intervention, we appended text (113 words) to the AI paragraph generation prompt that described results from Study~1 -- that AI writing assistance has a polarising effect; that the majority of writers dislike this effect -- and then asked the AI not to distort the writer's issue stance.
For the full prompt, see \textcolor{darkblue}{SM:\ref*{sm-subsec:mitigation_study_prompting_design}}.

For the Reranking intervention, we fine-tuned two reward models (RMs) on data collected in Study~1.
A first Paragraph RM, \(f_{\text{para}}\), was fine-tuned on paragraph-rating pairs (where ratings were averaged across readers at the paragraph level), to predict average third-party perception of new paragraphs.
A second Bullet RM, \(f_{\text{bullet}}\), was fine-tuned on bullet-rating pairs, i.e.\ bullet points provided by writers and average third-party ratings for the corresponding writer paragraphs.
During the study, at paragraph generation time, we used verbalised sampling~\citep{zhang2026verbalized} to generate two sets of five AI paragraph candidates, \(y_1,\dots,y_{k}\), from the assigned AI model.
Then, simultaneously, we used the Paragraph RM to predict the perceived issue stance of each AI-written candidate, \(s_k = f_{\text{para}}(y_k)\), and the Bullet RM to predict the perceived issue stance of the human-written paragraph based on the writer's bullet points, \(s^{\text{target}} = f_{\text{bullet}}(b)\).
Finally, we selected the AI-written paragraph candidate that minimised stance discrepancy, with directional weighting: \(y^* = y_{\arg\min_k \; w(\delta_k) \cdot \lvert \delta_k \rvert}\), where \(\delta_k = \lvert f_{\text{para}}(y_k) - 50 \rvert - \lvert f_{\text{bullet}}(b) - 50 \rvert\) measures how much more extreme (\(\delta_k > 0\)) or more moderate (\(\delta_k \leq 0\)) candidate \(k\) is predicted to appear relative to the writer's own text, and \(w(\delta) = 0.688\) if \(\delta > 0\) and \(w(\delta) = 0.448\) otherwise, so that polarising distortions are penalised more heavily than moderating ones, in proportion to the share of writers in Study~1 who objected to each direction of distortion (stated tolerance <50).

We tested multiple base models and training data sizes for the two RMs, finding clear benefits to scaling both.
The best-performing configuration, which we deployed in the mitigation study, used GPT-4.1 as the base model, fine-tuned via the OpenAI API on all available Study~1 data (Paragraph RM:\ 6,756 paragraph-rating pairs; Bullet RM:\ 3,378 bullet-rating pairs).
On a held-out test set, the two RMs achieved mean absolute errors of 6.15 and 9.28, respectively, in predicting perceived writer issue stance on a 0-100 scale.
For details, see \textcolor{darkblue}{SM:\ref*{sm-subsec:mitigation_study_reranking_design}}.

Statistical analysis followed the methods used in Study~1 .
To measure distortions of reader perception (N=2,543 readers) under each intervention vs the No Intervention control, we fit the same generalised linear mixed-effects regressions as in Study~1 with an intervention condition regressor and, \added{as specified for Study~3}, reader random effects \added{only} to each of the 29 rating attributes.
For full specifications and regression outputs, see \textcolor{darkblue}{SM:\ref*{sm-subsec:mitigation_study_distortions}}.
To quantify writer preference for AI writing under each intervention, we measured edit and preference rates with bootstrapped CIs, and fit logistic regressions with an intervention condition regressor and writer random effects to each binary outcome (\textcolor{darkblue}{SM:\ref*{sm-subsec:mitigation_study_writer_preference}}).

\subsection*{Trust study}

In Study~4 (N=802 readers), we investigated whether AI writing assistance has an effect on interpersonal trust placed in writers, and to what extent this effect varies across different levels of distortion from AI.
To obtain a behavioural measure of trust, we had readers play a Trust Game~\citep{berg1995trust}.
In every round of this Game, readers were shown a paragraph which they were told was written by another study participant (the ``author'').
Each reader received an endowment of 20p and could decide freely how much of this endowment to allocate to the author.
The author, readers were told, would receive 3$\times$ the allocated amount and could then decide how much to return.
Readers would be paid whatever portion of the endowment they kept plus whatever the author returned to them.
In practice, returns were fixed:\ at the end of the study, readers received the amount they kept plus 1.5$\times$ their total allocation across all paragraphs as a bonus payment.
We chose the multiplier such that, in comparison to returns based on live decisions in other studies, participants were better off on average~\citep{johnson2011trust}.
For details on Trust Game implementation and participant instructions, see \textcolor{darkblue}{SM:\ref*{sm-subsec:study_materials_trust_p2}}.

Each reader played four rounds of the Trust Game, with four paragraphs sampled from a subset of the paragraphs collected in Study~1.
To construct this subset, we began with all 1,501 paragraph pairs from Study~1 for which the writer's input to the AI was a set of bullet points.
Each pair consists of a human-written and an AI-assisted paragraph (including edits) from the same writer on the same proposition.
We removed paragraph pairs for which writers strictly preferred their own writing to the AI-assisted paragraph.
Then, we selected equally many ``low-distortion'' and ``high-distortion'' pairs based on the third-party ratings collected in Study~1:\
for each paragraph pair, we calculated distortion of a particular rating attribute (measured on a 0-100 scale, e.g.\ writer knowledge) as the difference between the rating of the AI-assisted paragraph and the rating of the corresponding human-written paragraph (where third-party ratings from Study~1 were averaged at the paragraph level).
We then took the average of absolute distortions across all 20 rating attributes that were measured on a 0-100 scale, and selected the 200 pairs with the highest as well as the 200 pairs with the lowest average, under constraint of equal selection across the three models that produced the AI-assisted paragraphs, so that the producing model does not confound the distortion contrast.
This resulted in 800 paragraphs, with 200 in each of four cells:\ paragraph type (human / AI) $\times$ amount of distortion (low / high).
Paragraph assignment was such that each Study~4 reader was assigned one paragraph from each cell in random order in a within-subjects design, without ever being assigned more than one paragraph per proposition.

The primary outcome variable was the reader's allocation amount (0-20p) per paragraph, and statistical analysis followed the methods used in Study~1 applied to this outcome.
We fit separate mixed-effects beta regressions, first with a single regressor for binary paragraph type (human / AI) to test whether readers placed more trust in AI-assisted than unassisted writers, then with an interaction between paragraph type and binary amount of distortion (low / high) to test whether the AI trust advantage was larger when distortion from AI was more pronounced.
All models included reader random effects.
For full specifications and regression outputs, see \textcolor{darkblue}{SM:\ref*{sm-subsec:trust_study_statistical_analysis}}.

\subsection*{Persuasion study}

In Study~5 (N=7,996 readers), we investigated whether AI writing assistance makes writing more persuasive, and to what extent this effect varies across different levels of distortion from AI.
To measure persuasion, we recorded reader stance on political propositions before and after reading an assigned paragraph on a 0-100 scale using a three-item battery~\citep{hackenburg2025levers} (see \textcolor{darkblue}{SM:\ref*{sm-subsec:study_materials_persuasion_p2}} for item wording).

Each reader read five paragraphs and gave their pre/post stance on five propositions.
Four paragraphs were sampled from the same 800 paragraphs selected for Study~4 described above, split across the same four cells:\ paragraph type (human / AI) $\times$ amount of distortion (low / high).
One paragraph corresponded to a control condition and showed off-topic informational writing (e.g.\ about coral reefs, lighthouses).
Paragraph assignment was such that each reader was assigned one paragraph from each cell plus control in random order in a within-subjects design, without ever being assigned more than one paragraph per proposition.
For details on the control condition, see \textcolor{darkblue}{SM:\ref*{sm-subsec:persuasion_study_paragraph_selection}}.

The primary outcome variable was the signed change in reader stance from before to after reading the assigned paragraph, oriented towards the directional position expressed in the paragraph as measured by third-party paragraph ratings from Study~1 (see \textcolor{darkblue}{SM:\ref*{sm-subsec:persuasion_study_descriptive_results}} for details).
We fit separate mixed-effects linear regressions:\
first with a single ternary regressor (human / AI / control paragraph) to test whether AI writing was more persuasive than human writing, then with a five-way regressor (human-high / human-low / AI-high / AI-low / control) to test whether the AI advantage was larger when distortion from AI was more pronounced.
All models included reader and proposition random effects.
For full specifications and regression outputs, see \textcolor{darkblue}{SM:\ref*{sm-subsec:persuasion_study_statistical_analysis}}.

\newpage

\bibliographystyle{unsrt}
\bibliography{references}

\newpage

\section*{Acknowledgments}

The authors would like to thank Dirk Hovy and the MilaNLP Lab at Bocconi University as well as the UK AI Security Institute's Societal Impacts Team for early feedback on this work.

\section*{Funding}

This work was funded under the UK AI Security Institute's Systemic AI Safety Grants programme.

\section*{Author Contributions}

All authors (P.R., K.H., H.R.K., C.S.) jointly conceptualised the study and designed the experiments.
P.R.\ developed the study interface, managed data collection, and conducted all data analysis.
P.R.\ wrote the original draft with substantial input from K.H.\ on Introduction and Results, and from C.S.\ on Introduction and Discussion.
All authors reviewed and edited the final manuscript.

\section*{Competing Interests}

The authors declare no competing interests.

\section*{Data and Code Availability}

All code, data and supplementary materials are available at:\ \href{https://github.com/paul-rottger/ai-distortion}{github.com/paul-rottger/ai-distortion}

\end{document}